\documentclass[10pt,twocolumn,letterpaper]{article}
\ifdefined\pdfminorversion
\fi
\usepackage[T1]{fontenc}
\usepackage{times,amsmath,amssymb,amsthm,mathtools}
\usepackage{booktabs,graphicx,xcolor,microtype,enumitem}
\usepackage[format=plain,labelsep=period,font=small]{caption}
\usepackage[numbers,sort&compress]{natbib}
\usepackage[hyphens]{url}
\definecolor{linkblue}{rgb}{0.21,0.49,0.74}
\usepackage[breaklinks,colorlinks,allcolors=linkblue]{hyperref}
\hypersetup{
  pdftitle={NullTok: Empirical-Null Token Selection for Training-Free Video Understanding},
  pdfauthor={Zhang Ziyue, Junwei Jian, Baiyang Song, Yunshan Zhong, Jun Peng, Yiyi Zhou, Rongrong Ji}
}

\setlist[itemize]{noitemsep,leftmargin=*,topsep=0pt}

\makeatletter
\renewcommand{\@maketitle}{%
  \begin{center}
    {\Large\bfseries \@title\par}
    \vspace{12pt}
    {\large\begin{tabular}{c}\@author\end{tabular}\par}
  \end{center}
  \vspace{9pt}}
\renewcommand{\section}{\@startsection{section}{1}{0pt}
  {-10pt plus -2pt minus -2pt}{7pt}{\normalfont\large\bfseries}}
\renewcommand{\subsection}{\@startsection{subsection}{2}{0pt}
  {-8pt plus -2pt minus -2pt}{5pt}{\normalfont\fontsize{11}{13}\selectfont\bfseries}}
\makeatother
\renewenvironment{abstract}
  {\begin{center}\large\bfseries Abstract\end{center}\noindent\itshape\ignorespaces}
  {\par\vspace{7pt}}
\newtheorem{theorem}{Theorem}
\newcommand{\method}{\textsc{NullTok}}

\title{NullTok: Empirical-Null Token Selection for Training-Free Video Understanding}
\author{%
  Zhang Ziyue\textsuperscript{1}, Junwei Jian\textsuperscript{3},
  Baiyang Song\textsuperscript{1}, Yunshan Zhong\textsuperscript{2},\\[3pt]
  Jun Peng\textsuperscript{1}, Yiyi Zhou\textsuperscript{1},
  Rongrong Ji\textsuperscript{1,*}\\[6pt]
  {\normalsize \textsuperscript{1}Xiamen University
  \qquad \textsuperscript{2}Hainan University}\\
  {\normalsize \textsuperscript{3}University of Science and Technology Beijing}\\[3pt]
  {\small \textsuperscript{*}Corresponding author}}
\date{}
\begin{document}
\maketitle
\raggedbottom
\begin{abstract}
Long video understanding remains a daunting challenge for existing \emph{Multi-modal Large Language Models} (MLLMs), mainly due to the large number of visual tokens accumulated in long videos. Currently, frame selection and token selection are two main paradigms for addressing this issue, both aiming to select relevant visual information to form a compact input for MLLMs. A natural approach combines these two paradigms by selecting frames before pruning tokens, but noisy frame-level relevance estimates can cause irreversible evidence loss. In this paper, we study this problem from the perspective of assessing the spatiotemporal distribution of query specific evidence and propose a novel and training-free approach termed NULLTOK. For every patch across all frames, we construct an empirical-null distribution from 64 group-matched irrelevant control queries, capturing the scores expected in the absence of target-specific evidence. Target affinities are calibrated against these same-position references into right-tail Monte Carlo $p$-values, which are discrete super-uniform under a conditional exchangeability assumption. Calibrated screening and a high-recall semantic guard produce candidate tokens for deterministic visual de-duplication and joint spatiotemporal budget allocation.
To validate NullTok, we apply it to two representative MLLMs and conduct extensive experiments on a set of video benchmarks. The experimental results demonstrate its effectiveness under a bounded visual token budget. For example, it achieves 65.81\% and 68.85\% on Video-MME using frozen LLaVA-Video-7B and Qwen3-VL-8B backbones, respectively, at a cap of 64 full-frame equivalents (FFE), without subtitles or audio. These results outperform all evaluated baselines on Video-MME under the same budget for both backbones.
Code is available at {\urlstyle{same}\url{https://github.com/potato-kitty/NullTok}}.
\end{abstract}
\section{Introduction}
\label{sec:introduction}

Long video understanding requires compressing a large visual stream into the bounded input of a Video LLM. As the number of frames grows, the resulting visual tokens impose substantial context and memory demands. Existing approaches address this constraint through frame selection or visual token compression. Frame selection can be viewed as a constrained form of token pruning, in which all tokens of a frame are retained or discarded together. Both operations therefore allocate the same finite budget over spatiotemporal tokens, consistent with the common compression perspective emerging in recent work \citep{Hao_2025_ICCV,Du_2026_CVPR}. Under this view, the goal is to select the tokens that preserve useful evidence for video question answering.

A natural extension is to combine these strategies by retrieving question-relevant frames or clips and then compressing their visual tokens. APVR combines pivot-frame retrieval with attention-driven token selection \citep{Gao_APVR_2026_AAAI}, while MARC adopts a retrieve-then-compress strategy with distillation based on reinforcement learning \citep{wu2025marc}. This ordering makes fine-grained selection depend on the coverage of the initial retrieval. However, these approaches based on frame selection can assign a low score to a frame containing a small object, a brief action, or a subtle attribute needed for the answer. Once that frame is excluded, subsequent token compression has no access to its tokens. The resulting risk is an early loss of locally useful evidence.

To address this risk, we formulate input compression as selecting valuable tokens across the sampled video. We assess every token across the sampled video before selecting the final frames and tokens, allowing evidence from any sampled position to enter a shared candidate pool. The goal is to preserve question-relevant evidence, whether it is localized to a small object, spans an action across several frames, or is distributed throughout a long recording. Global assessment uses a frozen dense encoder followed by batched similarity computations, sharing the feature processing structure of retrieval-based preprocessing. Dense features and compatible control responses can be cached and reused across multiple questions about the same video, amortizing the preprocessing cost. The downstream Video LLM receives only the selected sparse tokens.

This global view presents three challenges. First, raw query-token affinities lack a consistent interpretation across videos and token positions. A broadly applicable phrase may score highly against many tokens even when they provide little evidence for the current question, while visual texture and local feature scales can also influence these scores. Interpreting a high target affinity therefore requires a reference for how the same visual coordinate responds to irrelevant but structurally comparable queries. Second, useful objects or context may have only a weak contrast against background responses, requiring a high-recall screening policy. Third, relevant candidates can remain visually redundant and temporally uneven. A useful selector must therefore calibrate local affinities, protect potentially important semantics, and coordinate temporal coverage with spatial detail under one budget.

\begin{figure*}[t]
  \centering
  \includegraphics[width=0.9\linewidth]{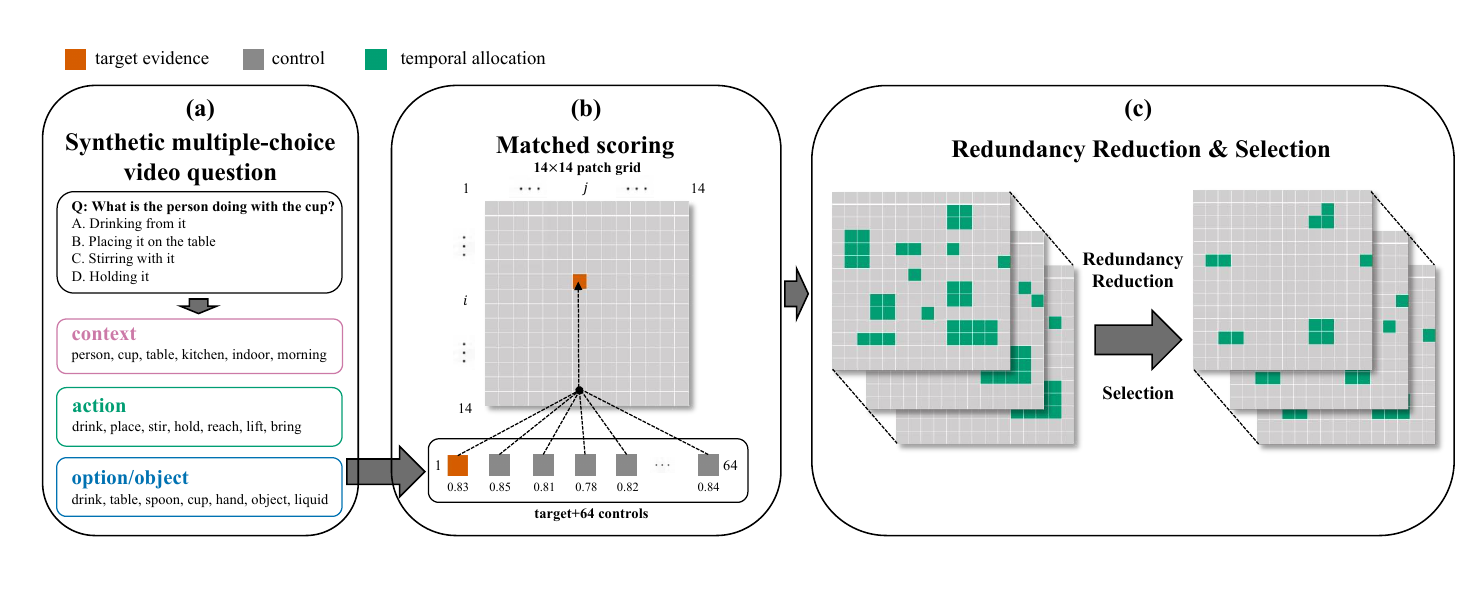}
  \caption{Method schematic of \method{}. Target and control queries are evaluated on identical video coordinates. Calibration precedes deterministic redundancy reduction and joint spatiotemporal selection. }
  \label{fig:pipeline}
\end{figure*}

We introduce \method, a training-free input compressor that constructs this reference directly from the target video. Control queries are evaluated on all sampled frames and token coordinates to characterize local background responses. For each target query, group-matched control scores form a local empirical null that converts the target affinity into a finite-sample right-tail Monte Carlo $p$-value. A small value indicates that few controls score as highly at that coordinate, providing a calibrated criterion for identifying potentially query-specific tokens. Combining this screen with a high-recall semantic guard produces the query-conditioned candidate set. Deterministic visual deduplication then reduces redundancy using frozen token embeddings, and a joint frame-token allocator distributes the budget according to the semantic relevance of the remaining candidates. The resulting input can preserve a few tokens from many temporal positions or denser evidence from fewer frames while leaving the downstream Video LLM unchanged. Figure~\ref{fig:pipeline} illustrates this sequence within one input budget.

We evaluate the proposed compressor on Video-MME, LongVideoBench, and MLVU with frozen LLaVA-Video-7B and Qwen3-VL-8B backbones \citep{Fu_2025_CVPR,wu2024longvideobench,Zhou_MLVU_2025_CVPR,zhang2024llavavideo,bai2025qwen3vl}. At a cap of 64 FFE, it achieves 65.81\% and 68.85\% on Video-MME, respectively. An additional comparison on Qwen-MLVU at 16 FFE examines lower-budget performance, while comparison against retrieval followed by compression examines the selection strategy. This cap specifies the total visual token budget, allowing the selector to distribute sparse tokens across a variable number of sampled frames while keeping the downstream model unchanged.

Our contributions are threefold:
\begin{itemize}[leftmargin=*,itemsep=1pt,topsep=2pt]
  \item We introduce same-video, same-position empirical-null calibration for
  query-conditioned token screening, with a finite-sample guarantee under
  conditional exchangeability and a clearly specified scope.
  \item We unify frame selection and token pruning within a deterministic
  spatiotemporal allocator that combines calibrated screening, semantic
  protection, and visual deduplication under one token budget.
  \item We evaluate the proposed compressor on three benchmarks and two
  backbones at 64 FFE, with an additional Qwen-MLVU comparison at 16 FFE,
  reporting both gains and limitations, including 65.81\% and 68.85\% on Video-MME at 64 FFE.
\end{itemize}

\section{Related Work}
\label{sec:related}

\paragraph{Video understanding benchmarks.}
Long-video benchmarks span different temporal scales and reasoning requirements. Video-MME covers short, medium, and long videos across six visual domains~\cite{Fu_2025_CVPR}; LongVideoBench studies referring reasoning with interleaved video and subtitles~\cite{wu2024longvideobench}; and LVBench targets extreme-duration comprehension~\cite{Wang_LVBench_2025_ICCV}. MLVU combines diverse genres and understanding tasks~\cite{Zhou_MLVU_2025_CVPR}, MVBench diagnoses dynamic and temporal understanding across 20 tasks~\cite{Li_MVBench_2024_CVPR}, and NExT-QA evaluates causal and temporal action reasoning~\cite{Xiao_NExTQA_2021_CVPR}. These differences motivate evaluation across benchmarks and backbones.

\paragraph{Query-aware temporal selection.}
Unlike uniform sampling, adaptive methods allocate temporal capacity using query relevance and video content. AKS balances relevance and coverage~\cite{Tang_2025_CVPR}, while Q-Frame coordinates frame selection and spatial resolution~\cite{Zhang_2025_ICCV}. QCA combines content-based segmentation, relevance allocation, and within-segment diversity~\cite{peng2026qca}; WFS uses wavelet-based semantic boundaries and diverse segment allocation~\cite{Chen_2026_CVPR}; and QuoTA decomposes queries for token assignment~\cite{luo2025quota}. Frame-level decisions can nevertheless discard locally useful tokens in low-scoring frames, motivating finer-grained evidence assessment before budget allocation.

\paragraph{Retrieval augmented and retrieve then compress pipelines.}
Video-RAG retrieves text from speech, optical character recognition, and object detection~\cite{Luo_VideoRAG_2025_NeurIPS}. APVR retrieves pivot frames before attention-driven token selection~\cite{Gao_APVR_2026_AAAI}, and MARC combines clip retrieval, compression, and reinforcement-learning-based distillation~\cite{wu2025marc}. ReKV uses query-conditioned KV-cache retrieval~\cite{Di_ReKV_2025_ICLR}; StreamKV couples semantic-segment retrieval with layer-adaptive compression~\cite{Chen_StreamKV_2026_AAAI}. Fine-grained selection in these pipelines is limited by initial retrieval coverage. Our QCA+MMTok baseline follows this ordering, whereas NULLTOK assesses all sampled token positions before candidate construction.

\paragraph{Visual token compression and unified accounting.}
FastV prunes tokens using early language-model attention~\cite{chen2024fastv}, while VisionZip retains dominant vision-encoder tokens and merges contextual features before language-model processing~\cite{yang2025visionzip}. DyTo combines hierarchical frame selection and dynamic token merging without task-specific training~\cite{zhang2024dyto}. LITE treats whole-frame selection as a limiting case of token allocation~\cite{Hao_2025_ICCV}, and Unified Spatiotemporal Token Compression uses a global retention pool~\cite{Du_2026_CVPR}. NULLTOK adopts this shared accounting unit; its contribution is the preceding assessment of whether local query affinity is exceptional relative to irrelevant queries.

\paragraph{Temporal redundancy and dynamic compression.}
DyCoke combines temporal token merging with dynamic KV-cache reduction~\cite{tao2025dycoke}. FrameFusion merges corresponding tokens across adjacent frames before importance-based pruning~\cite{fu2025framefusion}. These approaches distinguish importance from redundancy: highly attended tokens may still duplicate neighboring evidence.

\paragraph{Coverage and distribution-aware selection.}
MMTok jointly covers text tokens and the original visual-token set~\cite{dong2026mmtok}. OTPrune aligns retained and full token distributions through optimal transport~\cite{chen2026otprune}, while ReDiPrune combines text-conditioned relevance and max-min diversity~\cite{yu2026rediprune}. These complementary criteria motivate our token-pruning comparisons but differ from calibration against control queries at a fixed visual coordinate.

\paragraph{Score calibration and empirical nulls.}
Query-aware compressors commonly compare raw similarities, learned saliency, or within-video ranks.
PriorTR corrects visual token importance using an instruction-agnostic null-token attention prior and evaluates this strategy on video QA~\cite{chen2026priortr}.
Its reference is an attention prior obtained within the model, whereas NULLTOK constructs a distribution of responses to control queries using frozen semantic embeddings.
Empirical-null methodology is designed for settings in which a nominal theoretical null does not match the observed score distribution~\cite{Efron2004}.
Under exchangeability, Monte Carlo ranks give finite-sample tests even when the statistic is discrete or its analytic null is unavailable~\cite{Dufour2006}; the plus-one construction prevents a randomly sampled reference set from producing zero $p$-values and yields a conservative finite grid~\cite{PhipsonSmyth2010}.
NULLTOK transfers this principle to dense video evidence by holding the video, frame, and token coordinate fixed while replacing only the semantic query.
The finite-sample statement is tokenwise and marginal under conditional exchangeability; family-wise or false-discovery-rate control over the final mask would require additional analysis and is \emph{not} claimed.

\paragraph{Frozen Video LLMs and dense semantics.}
The compression setting considered here trains neither the Video LLM nor the semantic encoder.
LLaVA-Video and Qwen3-VL provide the frozen video-capable backbones used in the reported comparisons~\cite{zhang2024llavavideo,bai2025qwen3vl}.
DINO.txt aligns dense DINOv2 features with text and supplies spatially resolved semantic scores~\cite{Jose_2025_CVPR}.
NULLTOK uses these frozen representations to compare target and control phrases at identical positions.
All decisions are reproducible from the query, sampled features, empirical controls, and a fixed deterministic policy.

\paragraph{Positioning of NULLTOK.}
The closest conceptual precedents unify temporal and spatial compression or combine retrieval with token reduction, while the closest statistical precedents calibrate a target statistic against exchangeable controls.
NULLTOK combines these ideas in the following order: local null assessment at all sampled token positions $\rightarrow$ query-conditioned candidate screening $\rightarrow$ deterministic redundancy removal $\rightarrow$ joint frame-token allocation under one token budget.
Frozen visual features and compatible control responses can be cached and reused across questions; candidate sets and final masks remain query-conditioned.
This calibrated, training-free construction defines the scope of our contribution.

\section{Preliminaries}
\label{sec:preliminary}

\paragraph{Video understanding.}
Video understanding aims to extract semantic and temporal evidence from a video to answer queries about its content. We consider its query-conditioned form, video question answering (video QA). Given a video $V$ and a target query $b_0$ (which includes the question and answer options in a multiple-choice setting), a frozen Video LLM $f_\theta$ predicts an answer from a visual-token sequence. Let $\mathcal I=[T]\times[H]\times[W]$ index tokens from $T$ sampled frames, each with an $H\times W$ grid. For a retained index set $\mathcal S\subseteq\mathcal I$, the prediction is
\begin{equation}
  \hat y=f_\theta(b_0,\mathbf V_{\mathcal S}),
  \label{eq:video-qa-definition}
\end{equation}
where $\mathbf V_{\mathcal S}$ contains the model-native visual tokens $\xi_i$ at selected coordinates, serialized in temporal and spatial order. The full sampled input corresponds to $\mathcal S=\mathcal I$. Throughout this paper, assessment of all tokens refers to this sampled grid.

\paragraph{Frame and token selection.}
Input compression formulates video understanding as selecting a token subset $\mathcal S$ under a fixed visual token budget $K$, while preserving evidence needed to answer $b_0$. Existing approaches generally adopt two settings. Frame selection retains complete token grids at selected temporal positions. Formally, for a frame set $\mathcal F\subseteq[T]$, its output satisfies
\begin{equation}
  \mathcal S_{\mathrm{frame}}=\mathcal F\times[H]\times[W],
  \qquad |\mathcal F|HW\leq K.
  \label{eq:frame-selection-definition}
\end{equation}
Token selection operates at the level of individual tokens. Typically applied after initial frame retrieval $\mathcal F_0$, it chooses
\begin{equation}
  \mathcal S_{\mathrm{token}}\subseteq\mathcal F_0\times[H]\times[W],
  \qquad |\mathcal S_{\mathrm{token}}|\leq K.
  \label{eq:token-selection-definition}
\end{equation}
This frame-first setting permits spatial sparsity within the retrieved frames, with its temporal support rigidly fixed by $\mathcal F_0$.

\paragraph{Joint frame-token selection.}
Both paradigms allocate the same finite budget over spatiotemporal tokens, differing only in their unit of selection. Joint selection chooses $\mathcal S$ directly from $\mathcal I$. Writing $\mathcal S_t=\{(h,w):(t,h,w)\in\mathcal S\}$, its temporal and spatial allocation satisfies
\begin{equation}
  \mathcal F(\mathcal S)=\{t:|\mathcal S_t|>0\},
  \qquad |\mathcal S|=\sum_{t=1}^{T}|\mathcal S_t|\leq K.
  \label{eq:joint-selection-definition}
\end{equation}
Whole-frame selection is recovered as a special case when $|\mathcal S_t|\in\{0,HW\}$. Allowing intermediate counts supports allocating a few tokens from many temporal positions or denser evidence from fewer frames.

\paragraph{Query-specific evidence assessment.}
However, any budgeted selection requires a reliable criterion for the value of each visual coordinate, which the aforementioned definitions leave unspecified. Current paradigms suffer from two fundamental limitations. First, frame-first pipelines irreversibly discard locally useful evidence from the feasible pool. Second, raw query-token affinities lack a stable absolute interpretation across videos and spatial positions, conflating local visual salience with true query-specificity. To resolve this, we redefine input compression as a \emph{calibrated evidence distribution and budgeted allocation} problem. Instead of thresholding raw affinities, we assess whether a token's affinity is exceptionally high relative to a local reference distribution formed by structurally comparable control queries evaluated at the same coordinate. This provides a local reference for assessing query-specificity.

\section{Method}
\label{sec:method}

\subsection{Overview}
\label{sec:method-overview}

\begin{figure*}[t]
  \centering
  \includegraphics[width=0.9\linewidth]{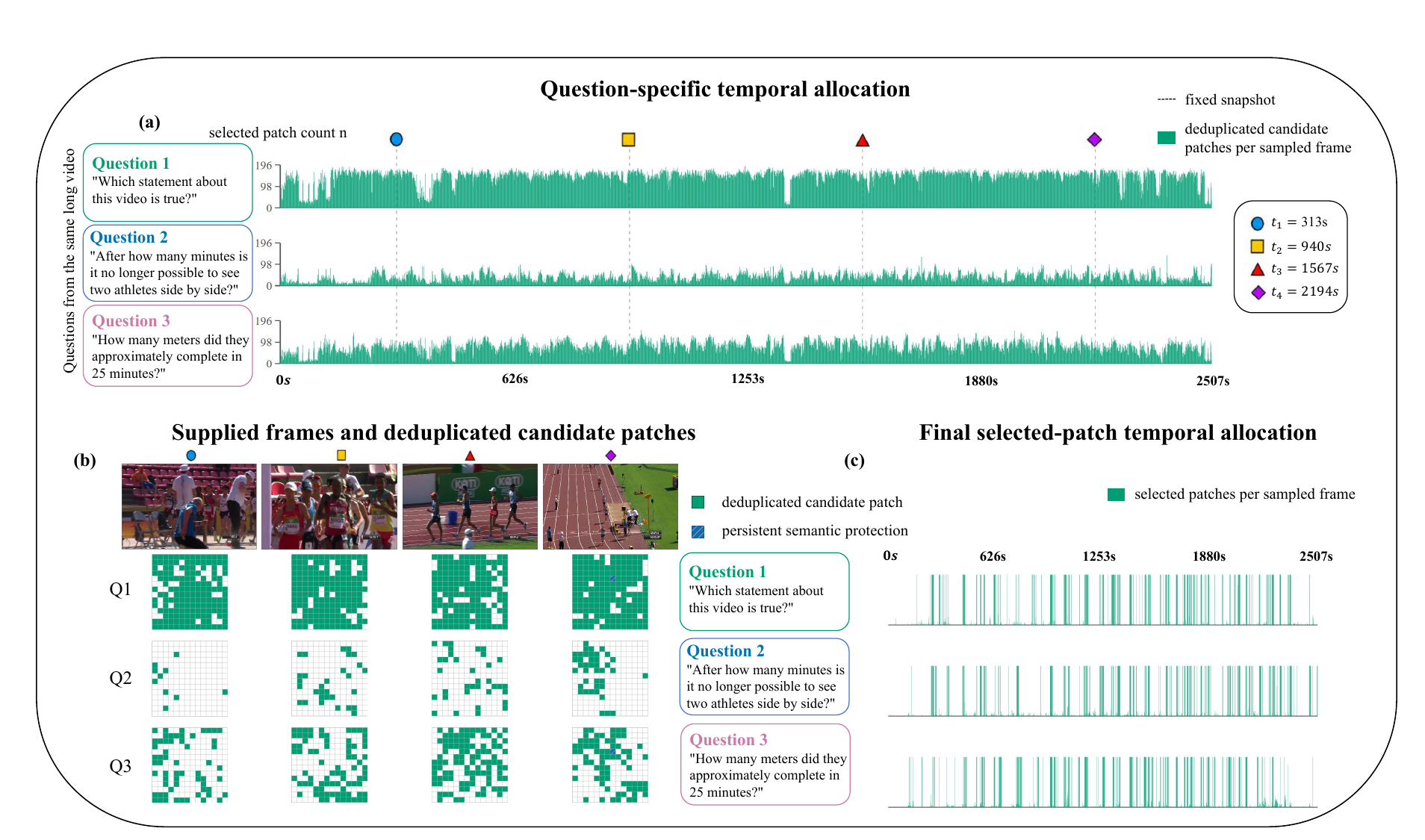}
  \caption{Query-dependent allocation for three questions from one
  preregistered long video (rows 2289--2291). (a) Deduplicated candidate-token
  counts per sampled frame. (b) Four fixed snapshots and their deduplicated
  candidate masks before final allocation; hatched tokens indicate persistent
  semantic protection. (c) Final selected-token counts per sampled frame.
  Each question retains exactly 12,544 tokens, spanning 617, 712, and 686
  nonempty frames for Q1, Q2, and Q3, respectively. The token budget is
  equivalent to 64 full $14\times14$ token grids, $64\times196=12{,}544$.}
  \label{fig:sparse-timeline}
\end{figure*}

Given $V$, $b_0$, and a token budget $K$, \method{} assesses all sampled tokens before budget allocation, as shown in Figure~\ref{fig:pipeline}. The question and options are decomposed into semantic groups (a), and the target and $B=64$ group-matched controls are scored at identical coordinates using frozen embeddings (b). Local empirical-null calibration and a high-recall semantic guard yield candidates $\mathcal C$. Deterministic visual deduplication produces per-frame sets $\mathcal D_t$, followed by joint frame-token allocation to select $\mathcal S\subseteq\bigcup_t\mathcal D_t$ with $|\mathcal S|\leq K$ (c). Selected model-native tokens are serialized for the unchanged Video LLM. Visual features and compatible control responses are reusable across questions when sampled coordinates and control-scoring configurations agree, while candidate sets and final masks remain query conditioned as shown in Figure~\ref{fig:sparse-timeline}.

\subsection{Query conditioned token statistics}

Using the sampled token index set $\mathcal I$ defined above, let $v_i$ be the frozen text-aligned visual embedding of token $i$. The question and options are decomposed into active semantic groups $\mathcal G\subseteq \{\mathrm{context},\mathrm{action},\mathrm{option}\}$. Let $\mathcal Q_{g b_j}$ be the filtered phrases in group $g\in\mathcal G$ for query $b_j\in\mathcal B=\{b_0,\ldots,b_B\}$, where $B$ denotes the number of control queries. Among them, $b_0$ is the target query relevant to the current video, while $b_1,\ldots,b_B$ are randomly sampled control queries. The per-token category winner and combined statistic are
\begin{align}
  s_{g b_j}(i) &= \max_{q\in\mathcal Q_{g b_j}}
  \cos\!\left(v_i,e(q)\right), \\
  Z_{b_j}(i) &= \sum_{g\in\mathcal G}\bar\lambda_g s_{g b_j}(i),
  &\bar\lambda_g&=\frac{\lambda_g}
  {\sum_{g'\in\mathcal G}\lambda_{g'}}.
  \label{eq:semantic-score}
\end{align}
Here $e(\cdot)$ is the frozen text encoder and $\lambda_g$ are fixed weights. Each control uses a complete phrase set from another video while its semantic score $Z_{b_j}(i)$ is calculated at the same token $i$. Therefore $(Z_{b_0}(i),\ldots,Z_{b_B}(i))$ contains the semantic scores of the same token under different queries. Target and control queries use the same active groups, phrase-filtering rule, aggregation, and weights.
The normalized target semantic score $M_i$ is calculated across sampled token positions as follows:
\begin{equation}
  M_i=\frac{Z_{b_0}(i)-\min_{k\in\mathcal I}Z_{b_0}(k)}
  {\max_{k\in\mathcal I}Z_{b_0}(k)-\min_{k\in\mathcal I}Z_{b_0}(k)+\varepsilon}.
  \label{eq:minmax}
\end{equation}

\subsection{Empirical-null calibration}

The finite-sample right-tail rank $p_i$ is
\begin{equation}
 \begin{aligned}
  p_i&=\frac{1+\sum_{j=1}^{B} 1\{Z_{b_j}(i)\geq Z_{b_0}(i)\}}{B+1},\\
   1\{Z_{b_j}(i)\geq Z_{b_0}(i)\} &=
   \begin{cases}
   1, & Z_{b_j}(i)\geq Z_{b_0}(i),\\
   0, & Z_{b_j}(i)<Z_{b_0}(i).
   \end{cases}
 \end{aligned}
  \label{eq:mc-pvalue}
\end{equation}
With screening level $\alpha$ and empirical quantile $Q_{.90}(M)$, the candidate mask is
\begin{align}
  \mathcal C_{\mathrm{MC}}&=\{i:p_i\leq\alpha\}, \\
  \mathcal C_{\mathrm{guard}}&=\{i:M_i\geq Q_{.90}(M)\}, \\
  \mathcal C&=\mathcal C_{\mathrm{MC}}\cup\mathcal C_{\mathrm{guard}}.
  \label{eq:guarded-candidates}
\end{align}
Figure~\ref{fig:pipeline}(b) illustrates matched scoring at one fixed visual coordinate.

\begin{theorem}[Conditional super-uniformity]
\label{thm:superuniform}
Fix $i$ and let $\mathcal H_i$ contain the video, sampled coordinates, active groups, and scoring rule. If $(Z_{b_0}(i),\ldots,Z_{b_B}(i))$ is conditionally exchangeable given $\mathcal H_i$, then for every $a\in[0,1]$,
\begin{equation}
  \Pr(p_i\leq a\mid\mathcal H_i)
  \leq\frac{\lfloor(B+1)a\rfloor}{B+1}\leq a.
  \label{eq:superuniform}
\end{equation}
\end{theorem}
The proof is provided in Appendix~\ref{app:proof}.

\subsection{Semantic seeds and visual deduplication}

Candidates are ordered lexicographically by
\begin{equation}
  \pi(i)=(-M_i,\ p_i,\ t_i,\ h_i,\ w_i).
  \label{eq:candidate-order}
\end{equation}
Here $M_i$ is the token's relevance score to the target question and options, normalized across the video. The value $p_i$ counts how many of the $B$ control queries from other videos score at least as highly as the target query at the same token, with one added to the count and division by $B+1$. The coordinates $(t_i,h_i,w_i)$ specify the sampled frame, token row, and token column.

A fixed prefix $\mathcal R_0$ of this ordering supplies semantic seeds, high-scoring tokens that serve as initial visual representatives. All seeds are retained during deduplication, preventing strong query evidence from being discarded solely because of visual similarity. A protected prefix $\mathcal P\subseteq\mathcal R_0$ additionally survives final budget allocation; seeds in $\mathcal R_0\setminus\mathcal P$ have no such final-retention guarantee. The frozen policy enables these prefixes for larger original-video frame counts and sets both to the empty set in the other configurations. Prefix sizes are fixed by that observable-length policy, not by answer correctness.
We remove visually redundant tokens from $\mathcal C$ by comparing each newly visited token with retained representative tokens. The graph $G_{10}=(\mathcal C,E_{10})$ connects each candidate to up to eight spatial neighbors in the same frame and to candidates at the same spatial coordinate in the previous and next sampled frames.
The two visual features serve different purposes. For semantic scoring, $v_i$ is the text-aligned token embedding from DINO.txt. For visual deduplication, the embedding $u_i$ is obtained by normalizing the model-native visual token $\xi_i$, which in the LLaVA implementation is produced by the frozen visual encoder, multimodal projector, and spatial pooling. The Qwen comparisons use a backbone-specific wrapper.
For each token $i$, $x_i\in\{0,1\}$ records its deduplication decision. Here $x_i=1$ means that token $i$ is retained, whereas $x_i=0$ means that it is discarded as visually redundant. $c_i$ denotes the index of the retained representative assigned to token $i$. When $i$ is retained, $c_i=i$; a discarded token inherits the index of its representative, whose embedding is $u_{c_i}$. Initialize $(x_i,c_i)=(1,i)$ for every seed $i\in\mathcal R_0$. Use $\mathcal R_0$ as the initial frontier of a synchronous breadth-first search (BFS).
At each BFS level, a newly reached token $i$ collects the representative indices carried by its neighbors in the preceding frontier, from which we select the one whose embedding is most similar to $u_i$:
\begin{align}
  \mathcal K_i&=\{c_j:(j,i)\in E_{10},\nonumber\\
  &\qquad j\text{ is in the previous frontier}\},
  \\
  c_i^*&=\arg\max_{k\in\mathcal K_i}\cos(u_i,u_k).
  \label{eq:visual-center}
\end{align}
If multiple representatives have the same similarity, choose the first in ascending lexicographic coordinate order $(t,h,w)$. The index $c_i^*$ identifies the most similar available representative. Its similarity to token $i$ determines whether $i$ is redundant. If the similarity reaches the threshold $\tau$, discard $i$ and assign it to $c_i^*$. Otherwise, token $i$ is retained as a new representative:

\begin{equation}
  (x_i,c_i)=
  \begin{cases}
    (0,c_i^*), & \cos(u_i,u_{c_i^*})\geq\tau,\\
    (1,i), & \cos(u_i,u_{c_i^*})<\tau.
  \end{cases}
  \label{eq:visual-redundancy}
\end{equation}

In both cases, token $i$ enters the next BFS frontier and carries its representative index $c_i$. Thus, even a discarded token passes its retained representative to the next level, allowing the search to continue through redundant regions.

If the resulting set underfills the budget, deduplication is rerun from the original candidate set \(\mathcal C\) with progressively higher similarity thresholds, so that fewer tokens are removed as redundant.

\subsection{Unified spatiotemporal allocation}

Let $\mathcal D_t$ be the deduplicated candidate tokens in frame $t$, with $n_t=|\mathcal D_t|$. We assign each frame a relevance score
\begin{equation}
  A_t(\beta)=\frac{\sum_{i\in\mathcal D_t}M_i}
  {\max(n_t,1)^{\beta}}
  \label{eq:frame-mass}
\end{equation}
which balances the total query relevance in the frame against the number of remaining candidate tokens. Frames are ranked by decreasing $A_t(\beta)$, with $n_t$ and temporal order used as deterministic tie-breakers. Within each frame, candidate tokens are ordered according to Eq.~(\ref{eq:candidate-order}).
Let $\mathcal C$ denote the set of all candidate spatiotemporal tokens, and
let $\mathcal P$ be the protected seed subset defined above. It is chosen
from $\mathcal C$ by the same semantic ordering and survives deduplication.
Initializing the selected mask with $\mathcal S=\mathcal P$ reserves these
high-scoring tokens even when aggregate frame scores or per-frame allocation
would favor other candidates. These mandatory tokens consume $|\mathcal P|$
of the same budget $K$ and count toward their frames' cumulative targets;
they are not additional tokens outside the budget. Protection is disabled
when $\mathcal P=\varnothing$. It is a deterministic recall safeguard, distinct
from the P90 candidate guard and outside the theorem's statistical guarantee.
Throughout allocation,
\begin{equation}
  \mathcal P\subseteq\mathcal S\subseteq\bigcup_t\mathcal D_t\subseteq\mathcal C,
  \qquad |\mathcal S|\leq K.
  \label{eq:budget-invariants}
\end{equation}
Allocation proceeds in repeated passes over the ranked frames. At pass $r\geq1$, let $\mathcal S_{t-}^{(r)}$ denote the selected mask immediately before visiting frame $t$. The cumulative target number of tokens for frame $t$ is
\begin{align}
  m_t^{(r)}&=\min(n_t,rQ),\\
  \Delta_t^{(r)}&=\min\!\Bigl(
  \max(m_t^{(r)}-|\mathcal S_{t-}^{(r)}\cap\mathcal D_t|,0),\nonumber\\
  &\hspace{5em}K-|\mathcal S_{t-}^{(r)}|\Bigr).
  \label{eq:quantum}
\end{align}
\newcommand{\CurrentFFESixtyFourTable}{%
\begin{table}[t]
\centering
\small
\setlength{\tabcolsep}{4pt}
\renewcommand{\arraystretch}{1.05}
\caption{Accuracy (\%) at a cap of 64 Full Frame Equivalents (FFE). Bold and underlined numbers mark the best and second best scores within each backbone and benchmark.}
\label{tab:cross-benchmark-main}
\resizebox{\columnwidth}{!}{%
\begin{tabular}{lccc}
\toprule
Method & Video-MME & LongVideoBench & MLVU \\
\midrule
\multicolumn{4}{l}{\textit{LLaVA-Video-7B}} \\
\quad Uniform & 64.15 & 59.16 & 66.51 \\
\quad QCA & \underline{65.22} & 60.13 & \underline{68.97} \\
\quad AKS & 63.63 & 59.99 & 67.50 \\
\quad QuoTA & 63.30 & 58.56 & 68.42 \\
\quad WFS & 64.78 & \textbf{62.45} & \textbf{71.73} \\
\quad \textbf{NullTok} & \textbf{65.81} & \underline{60.66} & 67.73 \\
\addlinespace[3pt]
\multicolumn{4}{l}{\textit{Qwen3-VL-8B}} \\
\quad Uniform & 66.63 & 61.93 & 68.60 \\
\quad QCA & \underline{68.52} & \underline{65.89} & 72.19 \\
\quad AKS & 68.07 & 64.25 & 71.13 \\
\quad QuoTA & 67.37 & 63.28 & 70.30 \\
\quad WFS & 68.00 & \textbf{66.57} & \underline{74.82} \\
\quad \textbf{NullTok} & \textbf{68.85} & 64.40 & \textbf{74.86} \\
\bottomrule
\end{tabular}%
}
\end{table}
}

\newcommand{\CurrentFFESixteenFigure}{%
\begin{figure}[t]
\centering
\includegraphics[width=\columnwidth]{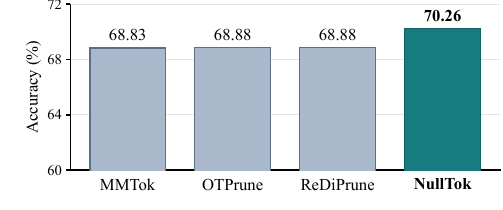}
\caption{MLVU accuracy on Qwen3-VL-8B at a cap of 16 Full Frame Equivalents (FFE). \method{} exceeds both OTPrune and ReDiPrune by 1.38\%. The vertical axis starts at 60\%.}
\label{fig:mlvu-ffe16}
\end{figure}
}

\newcommand{\CurrentVideoMMECombinedFigure}{%
\begin{figure}[t]
\centering
\includegraphics[width=\columnwidth]{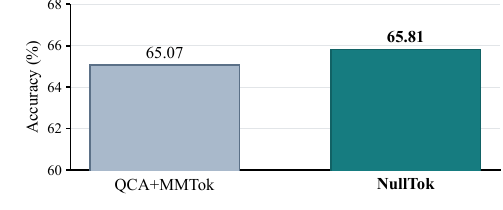}
\caption{Video-MME accuracy with LLaVA-Video-7B at 64 FFE. \method{} exceeds the combination of QCA and MMTok by 0.74\%. The vertical axis starts at 60\%.}
\label{fig:videomme-qca-mmtok}
\end{figure}
}
\newcommand{\CurrentNullAblationFigure}{%
\begin{figure}[t]
\centering
\includegraphics[width=\columnwidth]{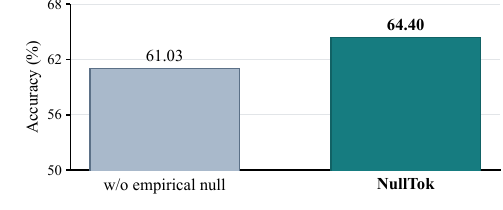}
\caption{Ablation results on the full LongVideoBench set, using Qwen3-VL-8B at 64 FFE. Replacing calibrated evidence with normalized semantic scores reduces accuracy by 3.37 \% points, demonstrating the effectiveness of our empirical-null token selection appraoch. The vertical axis starts at 50\%.}
\label{fig:null-ablation}
\end{figure}
}

\setcounter{topnumber}{1}
\setcounter{totalnumber}{1}
\CurrentFFESixtyFourTable
\CurrentVideoMMECombinedFigure
\CurrentFFESixteenFigure
\CurrentNullAblationFigure

Thus, each pass increases the per-frame target by at most $Q$ tokens, while $\Delta_t^{(r)}$ further limits the actual addition by the remaining global budget. The allocator then adds the next $\Delta_t^{(r)}$ ranked tokens from frame $t$. Passes continue until $|\mathcal S|=K$ or all candidates are exhausted.
The allocation behavior is controlled by three parameters: \(Q\) determines how many tokens a frame can acquire per pass, \(\beta\) controls the preference between total frame relevance and token density, and \(|\mathcal P|\) sets the amount of mandatory visual evidence preserved throughout allocation. Under a fixed global budget \(K\), varying these parameters continuously adjusts the temporal-spatial allocation pattern, ranging from whole-frame selection to sparse, frame adaptive token selection.

\subsection{Sparse serialization}

For each frame \(t\), the selected tokens are ordered in row-major order and
then serialized across frames in temporal order. The resulting visual sequence
is

\begin{equation}
  \mathbf V_{\mathcal S}
  =
  \underset{(t,h,w)\in\mathcal S}{\operatorname{Concat}}^{(t,h,w)\uparrow}
  \xi_{t,h,w},
  \label{eq:serialization}
\end{equation}

where \(\operatorname{Concat}\) preserves temporal order across frames and spatial order within each frame. This sparse sequence is directly fed into the unchanged QA model. 

\section{Experiments}
\label{sec:experiments}

\subsection{Experimental setup}

Our main experiments evaluate joint frame-token selection at a shared budget of 64 full-frame equivalents (FFE). We separately compare our framework with a pipeline combining frame selection and token pruning on Video-MME. Separately, we compare against token-pruning methods on MLVU with Qwen3-VL-8B at 16 FFE.

\paragraph{Experimental setting.}
We evaluate \method{} on three datasets, Video-MME~\citep{Fu_2025_CVPR}, LongVideoBench~\citep{wu2024longvideobench}, and MLVU~\citep{Zhou_MLVU_2025_CVPR}, using two frozen backbones, LLaVA-Video-7B~\citep{zhang2024llavavideo} and Qwen3-VL-8B~\citep{bai2025qwen3vl}. All experiments are conducted without subtitle inputs. A full-frame equivalent (FFE) measures the visual token budget in units of the number of tokens in one complete frame. For each baseline, we use its publicly released code and the parameter settings reported in its repository.

\paragraph{Comparisons.}
The compared methods are Uniform, which selects frames at evenly spaced temporal intervals throughout the video, QCA~\citep{peng2026qca}, AKS~\citep{Tang_2025_CVPR}, QuoTA~\citep{luo2025quota}, and WFS~\citep{Chen_2026_CVPR}. The token-pruning comparison is restricted to Qwen3-VL-8B on MLVU at 16 FFE and includes MMTok~\citep{dong2026mmtok}, OTPrune~\citep{chen2026otprune}, and ReDiPrune~\citep{yu2026rediprune}.

\subsection{Main results}

Table~\ref{tab:cross-benchmark-main} presents the main evaluation at 64 FFE. On Video-MME, \method{} achieves the highest accuracy with both backbones, 65.81\% on LLaVA and 68.85\% on Qwen. These results improve on Uniform by 1.66\% and 2.22\%, and surpass the strongest frame selection baseline, QCA, by 0.59\% and 0.33\%. Overall, \method{} achieves the best performance on Video-MME, the benchmark with the broadest coverage of video types. On LongVideoBench, \method{} reaches 60.66\% on LLaVA and 64.40\% on Qwen, exceeding Uniform by 1.50\% and 2.47\%. Although WFS remains stronger on both backbones and QCA is also stronger on Qwen. On MLVU, the Qwen result of 74.86\% improves on Uniform by 6.26\% and QCA by 2.67\%, while also higher than WFS (74.82\%).

\paragraph{Comparison against combined frame and token selection.}
To assess unified selection against sequential frame retrieval and token compression, we compare \method{} with QCA+MMTok on Video-MME using LLaVA-Video-7B. QCA first retrieves 256 frames, and MMTok compresses their visual tokens to the same 64-FFE budget. As shown in Figure~\ref{fig:videomme-qca-mmtok}, \method{} achieves 65.81\% accuracy compared with 65.07\% for QCA+MMTok, a gain of 0.74\%. This result shows that valuable information might be discard in frame-first stage of such combined pipeline. Instead, \method{} evaluates sampled tokens before budget allocation, allowing them to compete within a shared budget while balancing temporal coverage and spatial detail.

\paragraph{Comparison at a smaller visual budget.}
As an additional setting, Figure~\ref{fig:mlvu-ffe16} compares MLVU accuracy at a reduced budget of 16 FFE on Qwen. \method{} reaches 70.26\% on MLVU, exceeding MMTok by 1.43\% and both OTPrune and ReDiPrune by 1.38\%. These results demonstrate that \method{} also performs effectively for token pruning, outperforming methods specifically designed for token pruning under the same visual budget.

\paragraph{Ablations.}
To examine the role of calibration, we replace the empirical-null test with $M_i>1-\alpha$, using the video wide normalized semantic score in Eq.~\ref{eq:minmax}, where $\alpha$ is set to 0.3 for videos with at most 5,000 frames and 0.2 otherwise. All other settings remains unchanged. On all LongVideoBench questions, by doing so accuracy falls from 64.40\% to 61.03\% accoridng to Figure~\ref{fig:null-ablation}. This result demonstrates that our empirical-null token selection approach is effective.

\section{Conclusion}
\label{sec:conclusion}

We present \method{}, a training-free framework that treats frame selection and token pruning as a unified problem of evidence assessment and budget allocation. By calibrating query-token affinities against same-position empirical controls before selection, the framework allows evidence from all tokens to compete within a shared visual budget. Deterministic redundancy reduction and joint spatiotemporal allocation then turn these local assessments into a compact input for a frozen Video LLM. The main experiments demonstrate the promise of this formulation. On Video-MME, \method{} achieves 65.81\% and 68.85\% with LLaVA-Video-7B and Qwen3-VL-8B, which it the best reported results on through our experiments, and outperforms the combination of frame selection and token-purne, which is QCA + MMTok. Overall, \method{} provides a unified approach to visual token selection that jointly considers temporal coverage and spatial detail under a fixed budget. By grounding token selection in an empirical null model, \method{} provides a principled criterion for assessing selection evidence and achieves effective visual token allocation across different budget settings.

{\small
\bibliographystyle{ieeenat_fullname}
\bibliography{references,references_experiments}

\begin{thebibliography}{33}
\providecommand{\natexlab}[1]{#1}
\providecommand{\url}[1]{\texttt{#1}}
\expandafter\ifx\csname urlstyle\endcsname\relax
  \providecommand{\doi}[1]{doi: #1}\else
  \providecommand{\doi}{doi: \begingroup \urlstyle{rm}\Url}\fi

\bibitem[Bai et~al.(2025)Bai, Cai, Chen, et~al.]{bai2025qwen3vl}
Shuai Bai, Yuxuan Cai, Ruizhe Chen, et~al.
\newblock {Qwen3-VL} technical report.
\newblock \emph{arXiv preprint arXiv:2511.21631}, 2025.

\bibitem[Chen et~al.(2024)Chen, Zhao, Liu, Bai, Lin, Zhou, and
  Chang]{chen2024fastv}
Liang Chen, Haozhe Zhao, Tianyu Liu, Shuai Bai, Junyang Lin, Chang Zhou, and
  Baobao Chang.
\newblock An image is worth 1/2 tokens after layer 2: Plug-and-play inference
  acceleration for large vision-language models.
\newblock In \emph{Computer Vision -- ECCV 2024}, pages 19--35. Springer, 2024.

\bibitem[Chen et~al.(2026{\natexlab{a}})Chen, Zeng, Luo, Xie, Lin, Ji, Zhang,
  and Zheng]{Chen_2026_CVPR}
Wang Chen, Yuhui Zeng, Yongdong Luo, Tianyu Xie, Luojun Lin, Jiayi Ji, Yan
  Zhang, and Xiawu Zheng.
\newblock Wavelet-based frame selection by detecting semantic boundary for long
  video understanding.
\newblock In \emph{Proceedings of the IEEE/CVF Conference on Computer Vision
  and Pattern Recognition (CVPR)}, pages 24052--24061, 2026{\natexlab{a}}.

\bibitem[Chen et~al.(2026{\natexlab{b}})Chen, Zhu, Li, Dong, Xiong, Wang, Qiu,
  Song, Wang, Tang, Wang, and Razi]{chen2026otprune}
Xiwen Chen, Wenhui Zhu, Gen Li, Xuanzhao Dong, Yujian Xiong, Hao Wang, Peijie
  Qiu, Qingquan Song, Zhipeng Wang, Shao Tang, Yalin Wang, and Abolfazl Razi.
\newblock {OTPrune}: Distribution-aligned visual token pruning via optimal
  transport.
\newblock \emph{arXiv preprint arXiv:2602.20205}, 2026{\natexlab{b}}.

\bibitem[Chen et~al.(2026{\natexlab{c}})Chen, Bai, Wang, Bai, Dai, and
  Lu]{Chen_StreamKV_2026_AAAI}
Yilong Chen, Xiang Bai, Zhibin Wang, Chengyu Bai, Yuhan Dai, and Ming Lu.
\newblock {StreamKV}: Streaming video question-answering with segment-based
  {KV} cache retrieval and compression.
\newblock \emph{Proceedings of the AAAI Conference on Artificial Intelligence},
  40\penalty0 (4):\penalty0 3120--3128, 2026{\natexlab{c}}.

\bibitem[Chen et~al.(2026{\natexlab{d}})Chen, Cai, Guo, Cai, Yin, and
  Chen]{chen2026priortr}
Zengjie Chen, Yuxiang Cai, Jingcai Guo, Taotao Cai, Jianwei Yin, and Zhi Chen.
\newblock Accelerating multimodal large language models with prior-corrected
  token reduction.
\newblock In \emph{Computer Vision -- ECCV 2026}. Springer, 2026{\natexlab{d}}.

\bibitem[Di et~al.(2025)Di, Yu, Zhang, Li, Zhong, Cheng, Li, He, Shu, and
  Jiang]{Di_ReKV_2025_ICLR}
Shangzhe Di, Zhelun Yu, Guanghao Zhang, Haoyuan Li, Tao Zhong, Hao Cheng, Bolin
  Li, Wanggui He, Fangxun Shu, and Hao Jiang.
\newblock Streaming video question-answering with in-context video {KV}-cache
  retrieval.
\newblock In \emph{The Thirteenth International Conference on Learning
  Representations (ICLR)}, 2025.

\bibitem[Dong et~al.(2026)Dong, Hu, Zhang, Yin, Fu, and Qian]{dong2026mmtok}
Sixun Dong, Juhua Hu, Mian Zhang, Ming Yin, Yanjie Fu, and Qi Qian.
\newblock {MMTok}: Multimodal coverage maximization for efficient inference of
  {VLM}s.
\newblock In \emph{The Fourteenth International Conference on Learning
  Representations}, 2026.

\bibitem[Du et~al.(2026)Du, Xue, Li, Dai, and Lu]{Du_2026_CVPR}
Junhao Du, Jialong Xue, Anqi Li, Jincheng Dai, and Guo Lu.
\newblock Unified spatiotemporal token compression for video-{LLM}s at
  ultra-low retention.
\newblock In \emph{Proceedings of the IEEE/CVF Conference on Computer Vision
  and Pattern Recognition (CVPR)}, pages 17661--17671, 2026.

\bibitem[Dufour(2006)]{Dufour2006}
Jean-Marie Dufour.
\newblock Monte carlo tests with nuisance parameters: A general approach to
  finite-sample inference and nonstandard asymptotics.
\newblock \emph{Journal of Econometrics}, 133\penalty0 (2):\penalty0 443--477,
  2006.

\bibitem[Efron(2004)]{Efron2004}
Bradley Efron.
\newblock Large-scale simultaneous hypothesis testing: The choice of a null
  hypothesis.
\newblock \emph{Journal of the American Statistical Association}, 99\penalty0
  (465):\penalty0 96--104, 2004.

\bibitem[Fu et~al.(2025{\natexlab{a}})Fu, Dai, Luo, Li, Ren, Zhang, Wang, Zhou,
  Shen, Zhang, Chen, Li, Lin, Zhao, Li, Xu, Zheng, Chen, Shan, He, and
  Sun]{Fu_2025_CVPR}
Chaoyou Fu, Yuhan Dai, Yongdong Luo, Lei Li, Shuhuai Ren, Renrui Zhang, Zihan
  Wang, Chenyu Zhou, Yunhang Shen, Mengdan Zhang, Peixian Chen, Yanwei Li,
  Shaohui Lin, Sirui Zhao, Ke Li, Tong Xu, Xiawu Zheng, Enhong Chen, Caifeng
  Shan, Ran He, and Xing Sun.
\newblock {Video-MME}: The first-ever comprehensive evaluation benchmark of
  multi-modal {LLM}s in video analysis.
\newblock In \emph{Proceedings of the IEEE/CVF Conference on Computer Vision
  and Pattern Recognition (CVPR)}, pages 24108--24118, 2025{\natexlab{a}}.

\bibitem[Fu et~al.(2025{\natexlab{b}})Fu, Liu, Han, Dai, Yan, Yang, Ning, and
  Wang]{fu2025framefusion}
Tianyu Fu, Tengxuan Liu, Qinghao Han, Guohao Dai, Shengen Yan, Huazhong Yang,
  Xuefei Ning, and Yu Wang.
\newblock Framefusion: Combining similarity and importance for video token
  reduction on large vision language models.
\newblock In \emph{Proceedings of the IEEE/CVF International Conference on
  Computer Vision (ICCV)}, pages 22654--22663, 2025{\natexlab{b}}.

\bibitem[Gao et~al.(2026)Gao, Bao, Tu, Zhong, Yue, and
  Zhang]{Gao_APVR_2026_AAAI}
Hong Gao, Yiming Bao, Xuezhen Tu, Bin Zhong, Linan Yue, and Min-Ling Zhang.
\newblock {APVR}: Hour-level long video understanding with adaptive pivot
  visual information retrieval.
\newblock \emph{Proceedings of the AAAI Conference on Artificial Intelligence},
  40\penalty0 (6):\penalty0 4113--4121, 2026.

\bibitem[Hao et~al.(2025)Hao, Li, Gowda, Fisher, Huang, Arnab, and
  Sevilla-Lara]{Hao_2025_ICCV}
Xinyue Hao, Gen Li, Shreyank~N. Gowda, Robert~B. Fisher, Jonathan Huang, Anurag
  Arnab, and Laura Sevilla-Lara.
\newblock Principles of visual tokens for efficient video understanding.
\newblock In \emph{Proceedings of the IEEE/CVF International Conference on
  Computer Vision (ICCV)}, pages 21254--21264, 2025.

\bibitem[Jose et~al.(2025)Jose, Moutakanni, Kang, Baldassarre, Darcet, Xu, Li,
  Szafraniec, Ramamonjisoa, Oquab, Sim{\'e}oni, Vo, Labatut, and
  Bojanowski]{Jose_2025_CVPR}
Cijo Jose, Th{\'e}o Moutakanni, Dahyun Kang, Federico Baldassarre, Timoth{\'e}e
  Darcet, Hu Xu, Daniel Li, Marc Szafraniec, Micha{\"e}l Ramamonjisoa, Maxime
  Oquab, Oriane Sim{\'e}oni, Huy~V. Vo, Patrick Labatut, and Piotr Bojanowski.
\newblock {DINOv2} meets text: A unified framework for image- and pixel-level
  vision-language alignment.
\newblock In \emph{Proceedings of the IEEE/CVF Conference on Computer Vision
  and Pattern Recognition (CVPR)}, pages 24905--24916, 2025.

\bibitem[Li et~al.(2024)Li, Wang, He, Li, Wang, Liu, Wang, Xu, Chen, Luo, Wang,
  and Qiao]{Li_MVBench_2024_CVPR}
Kunchang Li, Yali Wang, Yinan He, Yizhuo Li, Yi Wang, Yi Liu, Zun Wang, Jilan
  Xu, Guo Chen, Ping Luo, Limin Wang, and Yu Qiao.
\newblock {MVBench}: A comprehensive multi-modal video understanding benchmark.
\newblock In \emph{Proceedings of the IEEE/CVF Conference on Computer Vision
  and Pattern Recognition (CVPR)}, pages 22195--22206, 2024.

\bibitem[Luo et~al.(2025)Luo, Zheng, Li, Yin, Lin, Fu, Huang, Ji, Chao, Luo,
  and Ji]{Luo_VideoRAG_2025_NeurIPS}
Yongdong Luo, Xiawu Zheng, Guilin Li, Shukang Yin, Haojia Lin, Chaoyou Fu,
  Jinfa Huang, Jiayi Ji, Fei Chao, Jiebo Luo, and Rongrong Ji.
\newblock {Video-RAG}: Visually-aligned retrieval-augmented long video
  comprehension.
\newblock In \emph{Advances in Neural Information Processing Systems}, 2025.

\bibitem[Luo et~al.(2026)Luo, Chen, Huang, Yin, Lin, Fu, Huang, Ji, Zheng, and
  Luo]{luo2025quota}
Yongdong Luo, Wang Chen, Weizhong Huang, Shukang Yin, Haojia Lin, Chaoyou Fu,
  Jinfa Huang, Jiayi Ji, Xiawu Zheng, and Jiebo Luo.
\newblock {QuoTA}: Query-oriented token assignment via {CoT} query decouple for
  long video comprehension.
\newblock \emph{Proceedings of the AAAI Conference on Artificial Intelligence},
  40\penalty0 (29):\penalty0 24160--24168, 2026.

\bibitem[Peng et~al.(2026)Peng, Song, Li, Li, Zhou, Ji, and Tian]{peng2026qca}
Jun Peng, Baiyang Song, Jie Li, Hui Li, Yiyi Zhou, Rongrong Ji, and Yonghong
  Tian.
\newblock {QCA}: Query- and content-aware keyframe selection for long video
  understanding.
\newblock In \emph{Computer Vision -- ECCV 2026}, pages 18--35. Springer, 2026.

\bibitem[Phipson and Smyth(2010)]{PhipsonSmyth2010}
Belinda Phipson and Gordon~K. Smyth.
\newblock Permutation p-values should never be zero: Calculating exact p-values
  when permutations are randomly drawn.
\newblock \emph{Statistical Applications in Genetics and Molecular Biology},
  9\penalty0 (1):\penalty0 Article 39, 2010.

\bibitem[Tang et~al.(2025)Tang, Qiu, Xie, Tian, Jiao, and Ye]{Tang_2025_CVPR}
Xi Tang, Jihao Qiu, Lingxi Xie, Yunjie Tian, Jianbin Jiao, and Qixiang Ye.
\newblock Adaptive keyframe sampling for long video understanding.
\newblock In \emph{Proceedings of the IEEE/CVF Conference on Computer Vision
  and Pattern Recognition (CVPR)}, pages 29118--29128, 2025.

\bibitem[Tao et~al.(2025)Tao, Qin, You, Sui, and Wang]{tao2025dycoke}
Keda Tao, Can Qin, Haoxuan You, Yang Sui, and Huan Wang.
\newblock Dycoke: Dynamic compression of tokens for fast video large language
  models.
\newblock In \emph{Proceedings of the IEEE/CVF Conference on Computer Vision
  and Pattern Recognition (CVPR)}, pages 18992--19001, 2025.

\bibitem[Wang et~al.(2025)Wang, He, Hong, Cheng, Zhang, Qi, Ding, Gu, Huang,
  Xu, Dong, and Tang]{Wang_LVBench_2025_ICCV}
Weihan Wang, Zehai He, Wenyi Hong, Yean Cheng, Xiaohan Zhang, Ji Qi, Ming Ding,
  Xiaotao Gu, Shiyu Huang, Bin Xu, Yuxiao Dong, and Jie Tang.
\newblock {LVBench}: An extreme long video understanding benchmark.
\newblock In \emph{Proceedings of the IEEE/CVF International Conference on
  Computer Vision (ICCV)}, pages 22958--22967, 2025.

\bibitem[Wu et~al.(2024)Wu, Li, Chen, and Li]{wu2024longvideobench}
Haoning Wu, Dongxu Li, Bei Chen, and Junnan Li.
\newblock {LongVideoBench}: A benchmark for long-context interleaved
  video-language understanding.
\newblock In \emph{Advances in Neural Information Processing Systems}, 2024.

\bibitem[Wu et~al.(2026)Wu, Yu, Liu, Wu, Zhou, and Shen]{wu2025marc}
Peiran Wu, Zhuorui Yu, Yunze Liu, Chi-Hao Wu, Enmin Zhou, and Junxiao Shen.
\newblock {MARC}: Memory-augmented {RL} token compression for efficient video
  understanding.
\newblock In \emph{The Fourteenth International Conference on Learning
  Representations (ICLR)}, 2026.

\bibitem[Xiao et~al.(2021)Xiao, Shang, Yao, and Chua]{Xiao_NExTQA_2021_CVPR}
Junbin Xiao, Xindi Shang, Angela Yao, and Tat-Seng Chua.
\newblock {NExT-QA}: Next phase of question-answering to explaining temporal
  actions.
\newblock In \emph{Proceedings of the IEEE/CVF Conference on Computer Vision
  and Pattern Recognition (CVPR)}, pages 9777--9786, 2021.

\bibitem[Yang et~al.(2025)Yang, Chen, Tian, Wang, Li, Yu, and
  Jia]{yang2025visionzip}
Senqiao Yang, Yukang Chen, Zhuotao Tian, Chengyao Wang, Jingyao Li, Bei Yu, and
  Jiaya Jia.
\newblock Visionzip: Longer is better but not necessary in vision language
  models.
\newblock In \emph{Proceedings of the IEEE/CVF Conference on Computer Vision
  and Pattern Recognition (CVPR)}, pages 19792--19802, 2025.

\bibitem[Yu et~al.(2026)Yu, Tsai, Zhang, Lu, Ye, and Chang]{yu2026rediprune}
An Yu, Ting~Yu Tsai, Zhenfei Zhang, Weiheng Lu, Felix X.-F. Ye, and Ming-Ching
  Chang.
\newblock {ReDiPrune}: Relevance-diversity pre-projection token pruning for
  efficient multimodal {LLM}s.
\newblock \emph{arXiv preprint arXiv:2603.24680}, 2026.

\bibitem[Zhang et~al.(2025{\natexlab{a}})Zhang, Yang, Yin, Luo, and
  Luan]{Zhang_2025_ICCV}
Shaojie Zhang, Jiahui Yang, Jianqin Yin, Zhenbo Luo, and Jian Luan.
\newblock {Q-Frame}: Query-aware frame selection and multi-resolution
  adaptation for video-{LLM}s.
\newblock In \emph{Proceedings of the IEEE/CVF International Conference on
  Computer Vision (ICCV)}, pages 22056--22065, 2025{\natexlab{a}}.

\bibitem[Zhang et~al.(2024)Zhang, Wu, Li, Li, Ma, Liu, and
  Li]{zhang2024llavavideo}
Yuanhan Zhang, Jinming Wu, Wei Li, Bo Li, Zejun Ma, Ziwei Liu, and Chunyuan Li.
\newblock Video instruction tuning with synthetic data.
\newblock \emph{arXiv preprint arXiv:2410.02713}, 2024.

\bibitem[Zhang et~al.(2025{\natexlab{b}})Zhang, Zhao, Chen, Ding, Yang, and
  Sun]{zhang2024dyto}
Yiming Zhang, Zhuokai Zhao, Zhaorun Chen, Zenghui Ding, Xianjun Yang, and
  Yining Sun.
\newblock Beyond training: Dynamic token merging for zero-shot video
  understanding.
\newblock In \emph{Proceedings of the IEEE/CVF International Conference on
  Computer Vision (ICCV)}, pages 22046--22055, 2025{\natexlab{b}}.

\bibitem[Zhou et~al.(2025)Zhou, Shu, Zhao, Wu, Liang, Xiao, Qin, Yang, Xiong,
  Zhang, Huang, and Liu]{Zhou_MLVU_2025_CVPR}
Junjie Zhou, Yan Shu, Bo Zhao, Boya Wu, Zhengyang Liang, Shitao Xiao, Minghao
  Qin, Xi Yang, Yongping Xiong, Bo Zhang, Tiejun Huang, and Zheng Liu.
\newblock {MLVU}: Benchmarking multi-task long video understanding.
\newblock In \emph{Proceedings of the IEEE/CVF Conference on Computer Vision
  and Pattern Recognition (CVPR)}, pages 13691--13701, 2025.

\end{thebibliography}
}
\newpage
\appendix
\section{Proof of conditional super-uniformity}
\label{app:proof}

Fix a token $i$. Let $Z_{b_0}(i)$ be its target-query score and
$Z_{b_1}(i),\ldots,Z_{b_B}(i)$ its control-query scores. Let $\mathcal H_i$
contain the video, sampled coordinates, active semantic groups, and scoring
rule. Assume these $B+1$ scores are conditionally exchangeable given
$\mathcal H_i$, and define
\[
 p_i=\frac{1+\sum_{j=1}^{B}1\{Z_{b_j}(i)\geq Z_{b_0}(i)\}}{B+1}.
\]
Condition on $\mathcal H_i$. First assume no ties. Exchangeability makes the
rank of $Z_{b_0}(i)$ among $(Z_{b_0}(i),\ldots,Z_{b_B}(i))$ uniform over
$\{1,\ldots,B+1\}$. The numerator of $p_i$ is its descending rank, so for
any $a\in[0,1]$,
\[
 \Pr(p_i\leq a\mid\mathcal H_i)
 =\frac{\lfloor(B+1)a\rfloor}{B+1}\leq a.
\]
With ties, counting every control satisfying $Z_{b_j}(i)\geq Z_{b_0}(i)$
assigns the target the least favorable descending rank in its tie block.
Its $p$-value is therefore no smaller than the normalized rank obtained
by breaking ties uniformly at random, where the rank is divided by $B+1$.
This preserves the inequality. Averaging over $\mathcal H_i$ gives the
unconditional form.

\end{document}